\documentclass[letterpaper, 10 pt, conference]{ieeeconf}  
\usepackage{array}
\usepackage{graphicx}
\usepackage{mathtools, amsmath, amssymb}
\usepackage{cite}

\IEEEoverridecommandlockouts                              

\title{\LARGE \bf
DGRC: An Effective Fine-tuning Framework for Distractor Generation in Chinese Multi-choice Reading Comprehension
}

\author{ Runfeng Lin$^{1}$, Dacheng Xu$^{2}$, Huijiang Wang$^{3}$, Zebiao Chen$^{4}$, Yating Wang$^{5}$ and Shouqiang Liu$^{*, 6}$ 
\thanks{*Corresponding author}
\thanks{$^{1, 2, 4, 6, 5}$School of Artificial Intelligence, Faculty of Engineering, South China Normal University,
    Nanhai, Foshan 528225, China 
    {\tt\small (runfenglin96, dacheng.xu, 2023025192, liusq, yatingwang)@m.scnu.edu.cn}}
\thanks{$^{3}$School of Computer Science and Engineering, Guangxi Normal University, Guilin 541004, China
    {\tt\small wanghuijiang111@126.com}}
}

\begin{document}

\maketitle
\thispagestyle{empty}
\pagestyle{empty}

\begin{abstract}

When evaluating a learner's knowledge proficiency, the multiple-choice question is an efficient and widely used format in standardized tests.
Nevertheless, generating these questions, particularly plausible distractors (incorrect options), poses a considerable challenge. 
Generally, the distractor generation can be classified into cloze-style distractor generation (CDG) and natural questions distractor generation (NQDG). 
In contrast to the CDG, utilizing pre-trained language models (PLMs) for NQDG presents three primary challenges:  
(1) PLMs are typically trained to generate ``correct'' content, like answers, while rarely trained to generate ``plausible" content, like distractors; 
(2) PLMs often struggle to produce content that aligns well with specific knowledge and the style of exams; 
(3) NQDG necessitates the model to produce longer, context-sensitive, and question-relevant distractors. 
In this study, we introduce a fine-tuning framework named DGRC for NQDG in Chinese multi-choice reading comprehension from authentic examinations. 
DGRC comprises three major components: hard chain-of-thought, multi-task learning, and generation mask patterns. 
The experiment results demonstrate that DGRC significantly enhances generation performance, achieving a more than 2.5-fold improvement in BLEU scores. 
\end{abstract}

\begin{figure*}[htbp]
 \centering
  \includegraphics[width=0.8\linewidth]{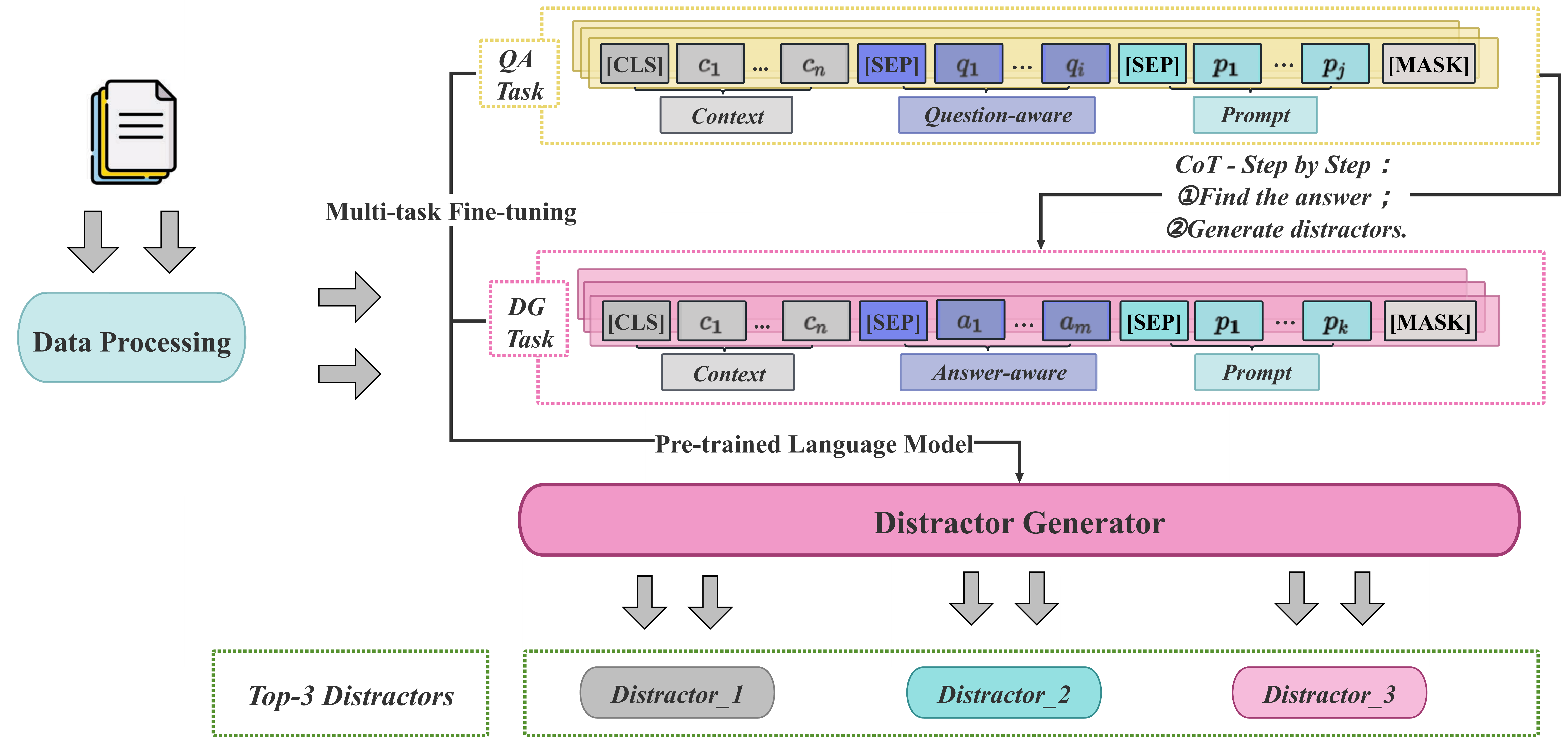}
\caption{Illustration of the framework: We employ multi-task learning and chain-of-thought (CoT) on the distractor generator approach in the distractor generator, which is based on pre-trained language models. Given the context, question, and answer, the distractor generator generates three distractors sequentially.}\label{fig:framework}
\end{figure*}
\section{INTRODUCTION}
Multi-choice testing (MCT) is widely recognized as one of the most enduring and successful forms of educational technology \cite{gierl2017developing, brown2001multi, ch2018automatic}. 
The effectiveness of an MCT relies greatly on the quality of its distractors. 
When distractors fail to confuse students, identifying the correct answer becomes too straightforward. 
Crafting functional and high-quality distractors, however, presents a significant challenge. 

Certain studies use similar words to generate distractors\cite{guo2016questimator, jiang2017distractor, brown2005automatic, chen2015interactive, ding2010automatic}. 
Concurrently, other works explore applying neural network-based methods in selecting distractors \cite{liang2017distractor, gao2019generating, zhou2020co, du2017learning}. 
Recent advancements in DG have witnessed a growing emphasis on leveraging PLMs to elevate the intricacies of generating contextually relevant distractors. 
Researchers, such as \cite{chung2020bert, offerijns2020better, kalpakchi2021bert, bitew2023distractor, peng2022misleading, chiang2022cdgp} have been actively engaged in the exploration of DG through the utilization of fine-tuning PLMs. 
Reference~\cite{chung2020bert} introduced multi-task learning and negative answer training strategies into DG, aiming to augment BERT's summarization ability. 
Their study delved into exploring both sequential and parallel approaches for multi-task learning. 
Additionally, they introduced the answer negative loss to mitigate BERT's tendency to predict the answers. 

However, the predominant research in DG primarily centers around cloze-style questions. 
These studies mainly involve generating words or phrases, relying on features extracted from datasets, including the position of the blank, part of speech, and sentence embedding where the blank is located. 

Typically, DG in cloze-style questions is easier than natural questions, where the question is absent, as in the cloze-style format, making it unnecessary for the generated distractors to relate to the question. 
Compared to cloze-style distractor generation, natural questions distractor generation (NQDG) with PLMs poses three main challenges: (1) PLMs are typically trained to generate ``correct" content, like answers to specific questions, while rarely trained to generate ``plausible" content like distractor. As a result, PLMs tend to generate answers even when tasked with DG; (2) PLMs often struggle to produce content that aligns well with specific knowledge and the style of exams; (3) NQDG demands a more intricate task from the model, involving not just simply synonym substitution but also natural language understanding and conditional generation, to produce longer, context-sensitive, and question-relevant distractors. 
Furthermore, few studies investigate NQDG in Chinese multi-choice reading comprehension due to the dataset's limited resources. 

To address those challenges, this paper introduces a novel fine-tuning framework for NQDG, encompassing elements such as hard chain-of-thought (hard CoT), multi-task learning, and generation mask patterns. 
We explore fine-tuning PLMs for DG across our compiled dataset, which includes two Chinese natural questions datasets derived from authentic examinations: $C^3$ \cite{sun2020investigating} and Logiqa \cite{liu2020logiqa}, with the exclusion of True/False questions and questions featuring fewer than four options. 
To guide the model in generating distractors that are contextually relevant to both the context and the question while ensuring it remains distinct from the answer, inspired by \cite{wei2022chain, lester2021power, dugan2022feasibility}, we incorporate hard CoT mechanism and multi-task learning into the distractor generator. 
Additionally, we structured our experiments around end-to-end and sequential mask patterns in multi-choice questions. 

To our knowledge, we are the first to venture into NQDG in Chinese multi-choice reading comprehension. 
Our principal contributions can be summarized as follows: 
\begin{itemize}
    \item We introduce a novel fine-tuning framework for natural questions distractor generation (DGRC), incorporating innovative elements such as hard chain-of-thought, multi-task learning, and generation mask patterns. Experiment results demonstrate that DGRC significantly enhances performance, achieving a more than 2.5-fold improvement in BLEU scores. 
    \item We propose a novel CoT approach, hard CoT, to make the input length acceptable and explicitly guide the model to reason, greatly improving the performance. 
    \item Due to the limited resources of Chinese multi-choice reading comprehension in natural questions, we compiled a composite dataset comprising \(C^3\) and Logiqa datasets with data cleaning.
\end{itemize}

\begin{figure*}[!htb]
  \centering
  \includegraphics[width=0.7\linewidth]{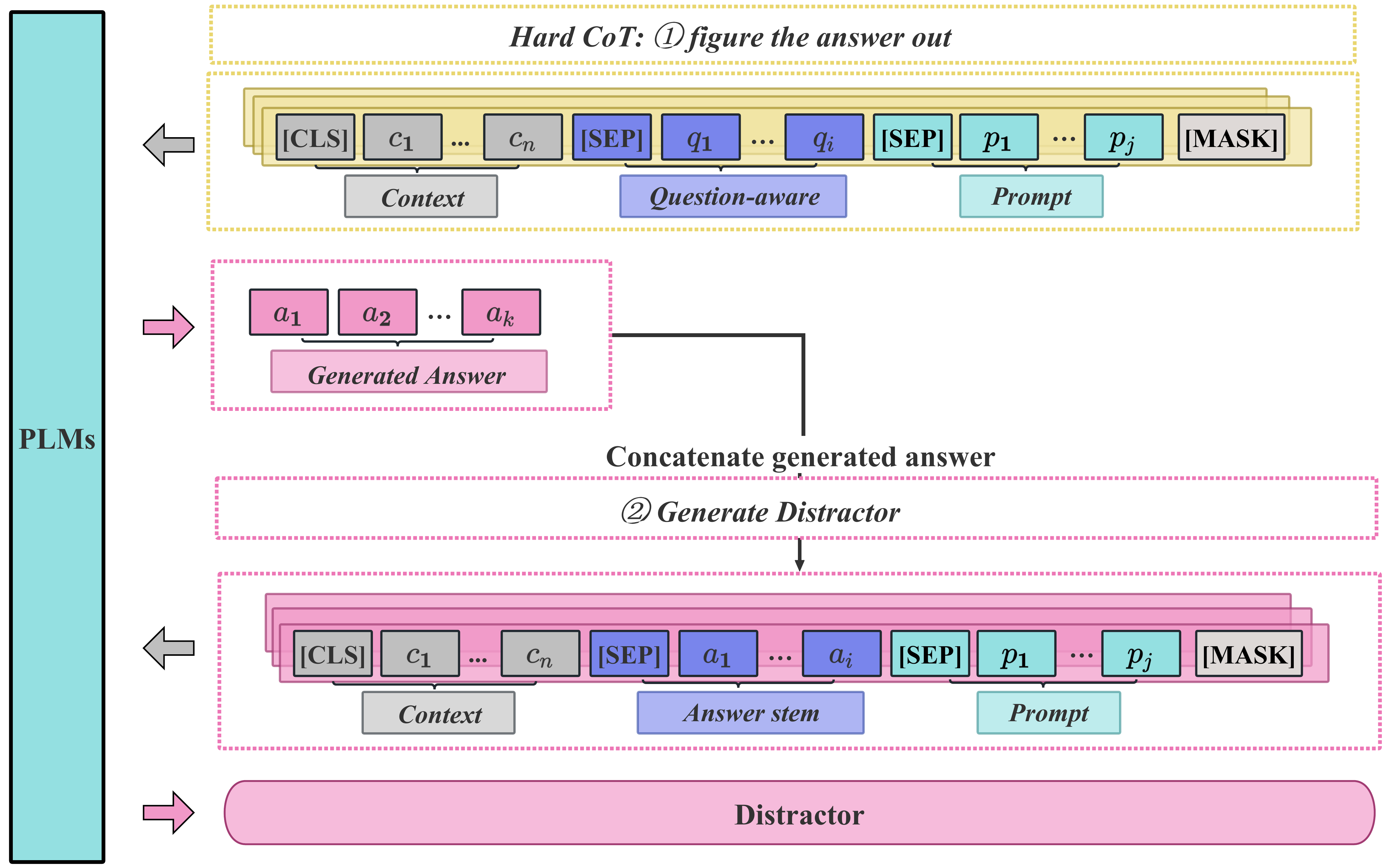}
  \caption{Illustration of hard CoT.}\label{fig:Hcot}
\end{figure*}

\section{Methodology}\label{sec:Methodology}
\subsection{Framework}\label{sec:framework}
DGRC comprises three major components: hard chain-of-thought, multi-task learning, and generation mask patterns, as illustrated in Fig.~\ref{fig:framework}. 

We define the following configurations to investigate the impact of distinct fine-tuning settings. 

\subsection{Fine-Tuning Strategies}
\subsubsection{Question-aware Fine-Tuning}\label{sec:question-aware ft}
During question-aware fine-tuning, we directly input the reading comprehension context \(C\) into PLM (denoted by \(\mathbb{M}\)), concatenating it with the question stem \(Q\), corresponding prompt \(P\), and [MASK] token (denoted by \(C_{\otimes (Q, P,\mathrm{[MASK]})}\)). 
Fine-tuning aims to identify a parameter set \(\theta\) that minimizes the loss function \(\log p\). 

\begin{equation}
    \mathbb{M}(C_{\otimes (Q,P,\mathrm{[MASK]})}) \to D
\end{equation}

\begin{equation}
    \mathop{minimize}\limits_{\theta} -\sum\limits_{\forall C} log(p(D|C,Q,P;\theta))
\end{equation}

\subsubsection{Answer-aware Fine-Tuning}\label{sec:answer-aware ft}
In answer-aware fine-tuning, we incorporate the answer stem \(A\) between \(C\) and \((P,\mathrm{[MASK]})\), separated by the \(\mathrm{[SEP]}\) token. 
Answer stem \(A\) is employed to guide the model in generating distractor \(D\). 
Specifically, 

\begin{equation}
    \mathbb{M}(C_{\otimes (A,P,\mathrm{[MASK])} }) \to D
\end{equation}

The training objective is to minimize the following loss function while seeking an optimal parameter set \(\theta\). 

\begin{equation}
    \mathop{minimize}\limits_{\theta} -\sum\limits_{\forall (C,A)} log(p(D|C,A,P;\theta)).
\end{equation}

\subsubsection{Question-enhanced Answer-aware Fine-Tuning}\label{sec:question-answer-aware ft}
In this fine-tuning approach, we propose the adoption of question-enhanced answer-awareness.  

\begin{equation}
    \mathbb{M}(S_{\otimes \mathrm{[MASK]} }) \to D
\end{equation}

The model input stem \(S\)  is composed of the context \(C\), question-aware stem \(Q\), answer-aware stem \(A\) , prompt \(P\), and \(\mathrm{[MASK]}\) token. 
When fine-tuning, the model predicts \(D\). 
The loss function is as follows, and the training objective is to identify a parameter set \(\theta\) that minimizes the loss. 

\begin{equation}
    \mathop{minimize}\limits_{\theta} -\sum\limits_{\forall S} log(p(D|C,Q,A,P;\theta))
\end{equation}

\subsection{Hard Chain-of-Thought}\label{sec:HCoT}
Few-shot chain-of-thought (few-shot CoT) has proven effective in investigations \cite{gramopadhye2024few, liang2023prompting, xu2023multi, peng2020few}. However, when it comes to generating reading comprehension distractors, it tends to create overly lengthy inputs for PLMs.
Additionally, fine-tuning PLMs with few-shot CoT may lead to uncertainty in adhering to its guiding principles. 
To address these issues, we propose a novel approach named hard chain-of-thought, illustrated in Fig.~\ref{fig:Hcot}. 

The hard CoT mechanism directs the model to prioritize deducing the answer before generating distractors. 
Moreover, introducing hard CoT reduces the model's reliance on the answer-aware stem.  
Models fine-tuned with hard CoT can generate distractors without depending on dataset answers. 
\begin{figure}[!htb]
  \centering
  \includegraphics[width=1\linewidth]{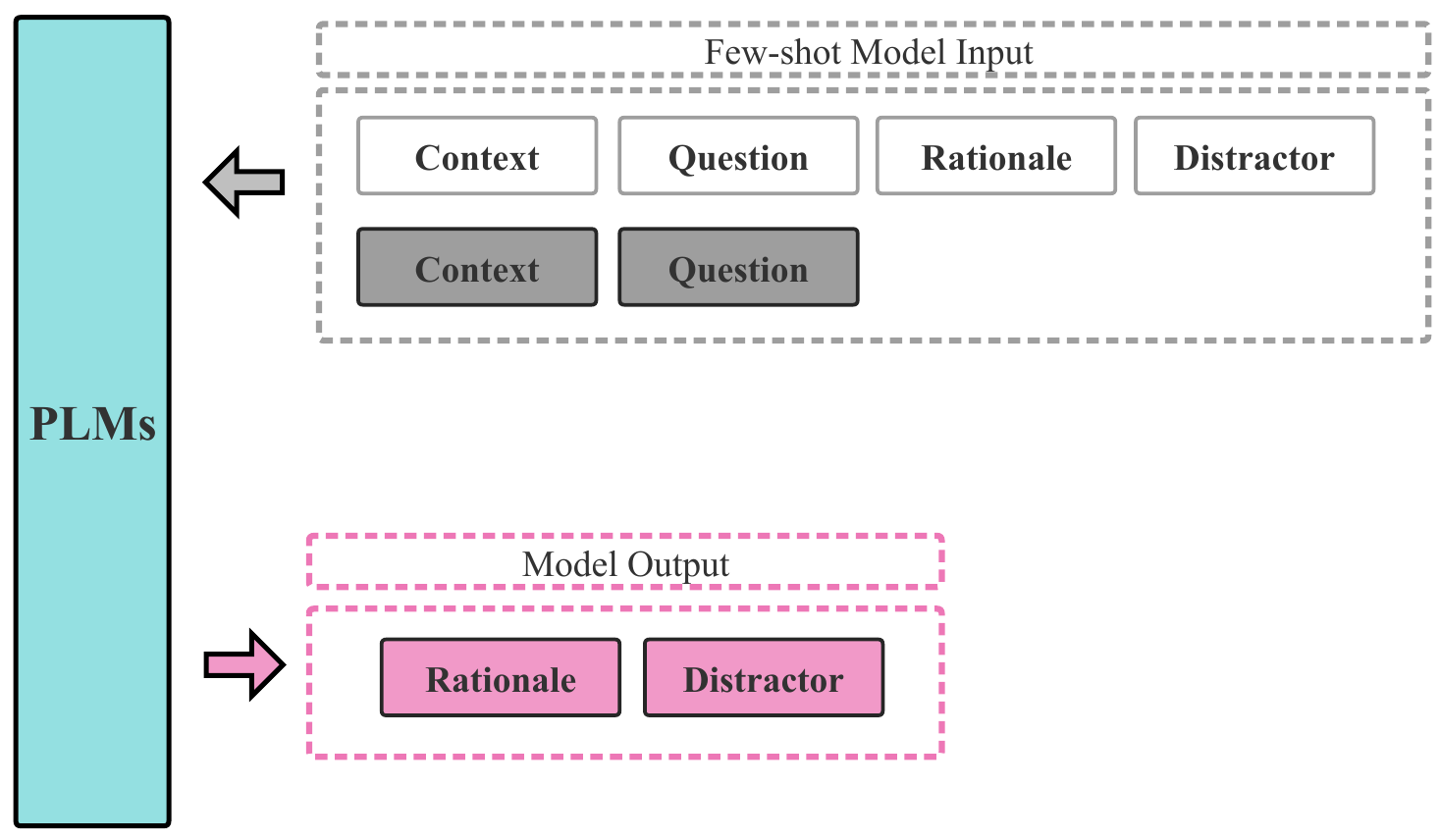}
  \caption{Illustration of few-shot CoT.}\label{fig:Few-shot CoT}
\end{figure}

\subsection{Hard CoT and Multi-task Learning Strategies}\label{sec:fine-tuning strategy}
We incorporate question-answering and distractor generation into multi-task learning.  
The formulations for hard CoT and multi-task learning are as follows:

\begin{equation}
    \mathop{minimize}\limits_{\theta} - \sum\limits_{\forall S} \phi_{ML}(S,C_{qa})\\
\end{equation}
\begin{equation}
    \phi_{ML}=\psi_{DG}(S)+\gamma\cdot\psi_{QA}(C_{qa})+ \delta\cdot\psi_{CoT}(C_{cot})\\
\end{equation}

The \(S\) stem and \(C_{qa}\) stem serve as inputs for distractor generation fine-tuning and question answering fine-tuning, respectively.  
The loss function aims to minimize \(\psi_{DG}(S)+\gamma\cdot\psi_{QA}(C_{qa})+ \delta\cdot\psi_{CoT}(C_{cot})\). 
Here,  \(\gamma\) and \(\delta\) function as hyper-parameters regulating weighting among the three tasks. 

\begin{equation}
    \psi_{DG}(S) = \sum\limits_{\forall S} \log(p(D|S;\theta)) \\
\end{equation}
\begin{equation}
     \psi_{QA}(C_{qa})=\sum\limits_{\forall {(C,Q)}} \log(p(D|C,Q,P_{qa};\theta)) \\
\end{equation}
\begin{equation}
    \psi_{CoT}(C_{cot}) = \sum\limits_{\forall {(C,Q)}} \log(p(D|C,Q,P_{cot};\theta))
\end{equation}

\begin{figure}[!htb]
  \centering
  \includegraphics[width=1\linewidth]{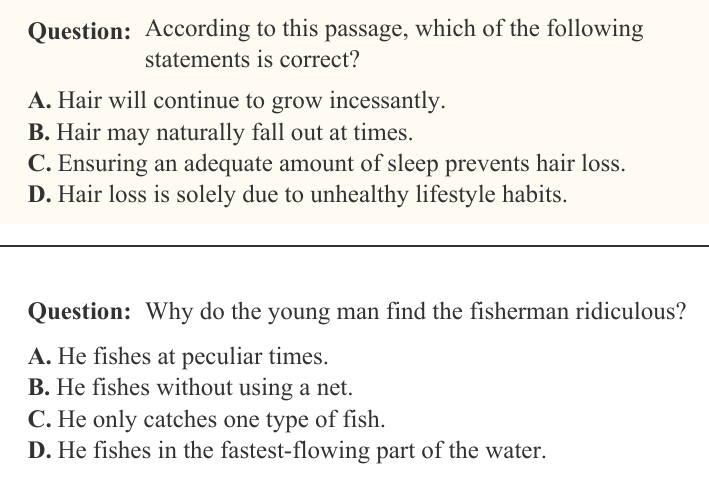}
  \caption{Templated question (above) and non-templated question (below).}\label{fig: templated and non-templated questions}
\end{figure}

We have classified the questions into templated and non-templated and devised distinct hard CoT and multi-task learning strategies, respectively. 
As depicted in Fig.~\ref{fig: templated and non-templated questions}. 

\subsubsection{Templated Questions}
\begin{figure}[!htb]
  \centering
  \includegraphics[width=1\linewidth]{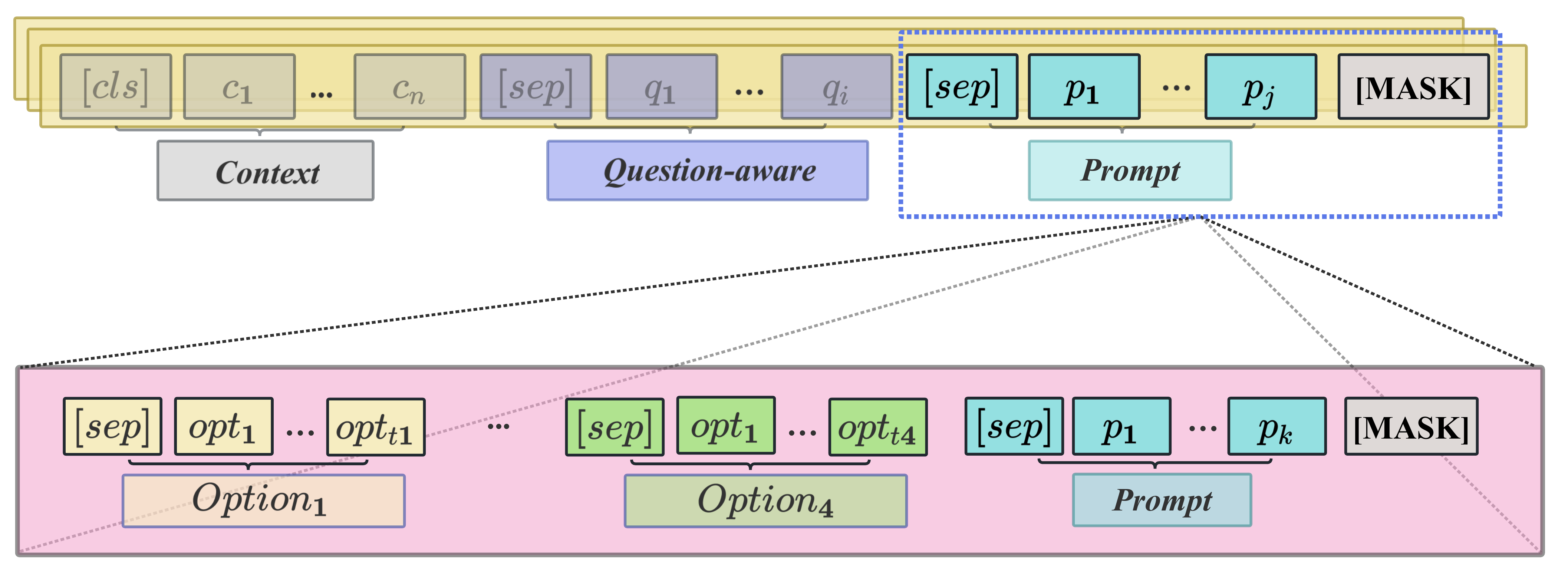}
  \caption{Formulation of multi-choice question answering task.}\label{fig: multi-choice question answering}
\end{figure}

Templated questions typically lack specific contextual content, themes, and information for the model to reason about. 
Fine-tuning directly conducting question-answering tasks in multi-task learning and hard CoT is ineffective for templated questions. 
Therefore, we exclude hard CoT fine-tuning and transform question answering into a multi-choice question answering task, as depicted in Fig.~\ref{fig: multi-choice question answering}. 

\subsubsection{Non-Templated Questions}
For non-templated questions, we utilize both hard CoT and multi-task learning. 

\subsection{Generation Mask Patterns}\label{sec:mask pattern}
\begin{figure}[!htb]
  \centering
  \includegraphics[width=1\linewidth]{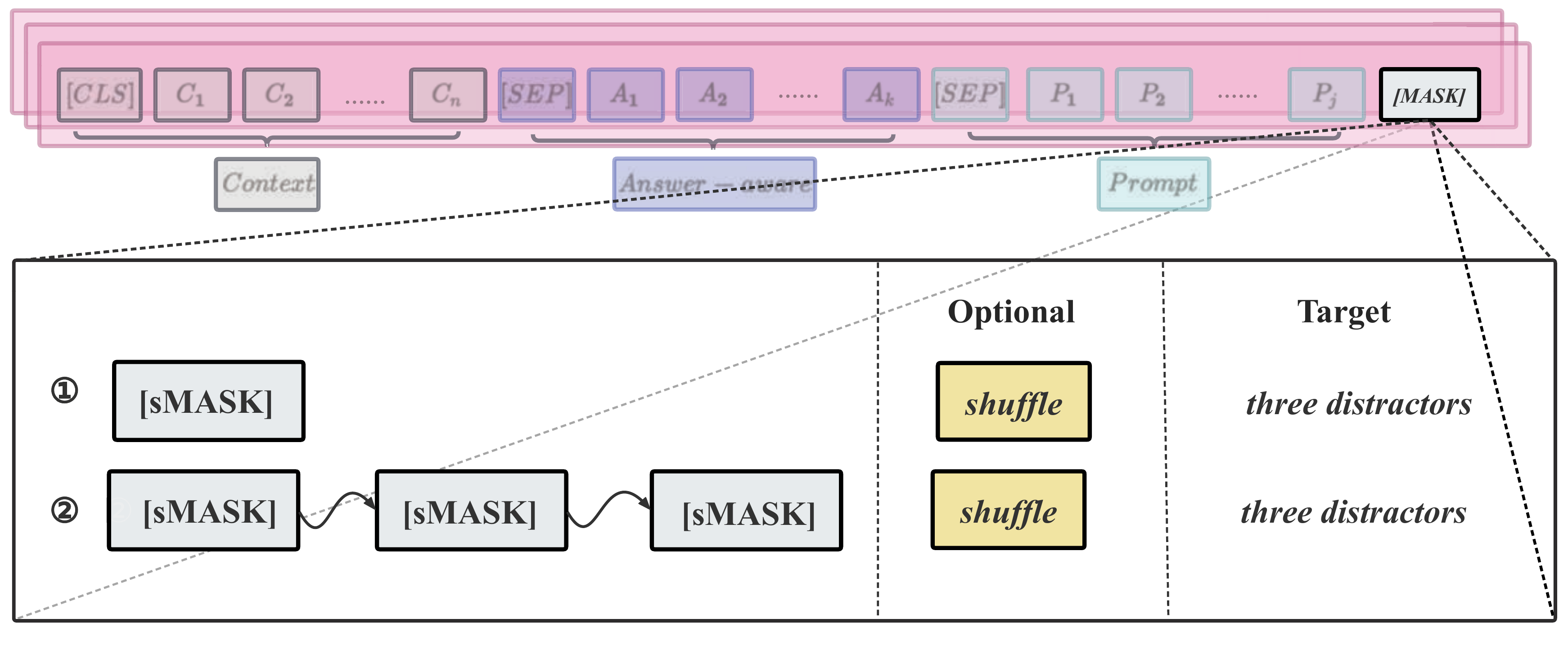}
  \caption{Illustration of generation mask patterns.}\label{fig:mask pattern}
\end{figure}

This section introduces the generation mask patterns, depicted in Fig.~\ref{fig:mask pattern}. 

\subsubsection{End-to-End Mask Pattern}
In pursuit of the end-to-end generation objective, we explore the end-to-end mask pattern in that the model simultaneously generates three distractors during fine-tuning. 
We employ one [MASK] token to generate three distractors. 

\subsubsection{Sequential Mask Pattern}
We introduce the sequential mask pattern to assess the influence of conditional generation.  
Sequential generation entails the model producing the first distractor and then learning to generate the second distractor based on the previously generated ones, continuing in this manner. 

Note that GLM \cite{du2021glm} is trained with distinct objectives and utilizes three types of [MASK] tokens, namely [MASK], [sMASK], and [gMASK]. 
We conduct our experiments on the [sMASK] token. 

To minimize the impact of the dependency on the labels' order, we also introduce a shuffle mechanism to randomize the sequence of labels during the fine-tuning process.  

\begin{table*}
    \centering
	\renewcommand\arraystretch{1.3}
    \caption{Automatic evaluation results on a mixed validation set of \(C^3\) and Logiqa}\label{tab:exp}
    \begin{tabular}{|l|c|c|c|c|c|c|}
      \hline
      \textbf{Model} & \textbf{BLEU-1}  & \textbf{BLEU-2}  & \textbf{BLEU-3}  & \textbf{BLEU-4}  & \textbf{METEOR}  & \textbf{ROUGE-L} \\
      \hline 
      GLM(Q-aware)$^{\mathrm{a}}$ & 36.73 & 21.61 & 13.00 & 7.49 & 3.72 & 4.32\\
      \hline
GLM(A-aware) & 39.62 & 23.89 & 14.80 & 8.81 & 6.61 & 7.91\\
\hline
      GLM(QA-aware) $^{\mathrm{a}}$& 40.12 & 24.31 & 15.14 & 9.06 & 7.52 & 8.60\\
\hline
      ChatGLM3-6B(Q-aware) $^{\mathrm{a}}$& 38.35 & 22.81 & 13.87 & 8.09 & 10.30 & 8.15 \\
      \hline
      ChatGLM3-6B(A-aware) $^{\mathrm{a}}$ & 37.79 & 22.45 & 13.66 & 8.18 & 9.88 & 8.03\\
      \hline
      ChatGLM3-6B(QA-aware)$^{\mathrm{a}}$ & 39.35 & 23.52 & 14.37 & 8.43 & 10.24 & 9.01\\
      \hline
      DGRC$_{(ft1,seq)}$ $^{\mathrm{b, c}}$& 49.77 & 30.76 & 19.42 & 11.83 & 18.86 & 25.42\\
      \hline
      DGRC$_{(ft1,seq,shuf)}$ $^{\mathrm{b, c}}$& 50.28 & 31.20 & 19.82 & 12.16 & 19.43 & 24.72 \\ 
      \hline
      DGRC$_{(ft1,e2e)}$ $^{\mathrm{b, c}}$&54.87 & 34.43 & 22.05 & 13.61 & 23.16 & 32.27 \\
      \hline
      DGRC$_{(ft1,e2e,shuf)}$ $^{\mathrm{b, c}}$& 54.43 & 34.24 & 21.98 & 13.59 & 24.04 & 29.87\\
      \hline
      DGRC$_{(ft2,seq)}$ $^{\mathrm{b, c}}$& 57.87 & 38.50 & 26.28 & 21.26 & 22.43 & 33.71\\
      \hline
      DGRC$_{(ft2,seq,shuf)}$ $^{\mathrm{b, c}}$& 57.77 & 37.39 & 24.82 & 17.30 & 22.26 & 32.65 \\
      \hline
      DGRC$_{(ft2,e2e)}$ $^{\mathrm{b, c}}$& 61.45 & 40.63 & 27.59 & \textbf{22.64} &  28.08 & 37.20\\
      \hline
      DGRC$_{(ft2,e2e,shuf)}$ $^{\mathrm{b, c}}$& 62.83 & 41.62 & 28.25 & 21.74  & 29.00 & 37.50\\ 
\hline
      DGRC$_{(ft3,seq)}$  $^{\mathrm{b, c}}$& 58.77 & 38.20 & 25.47 & 16.38 & 23.06 & 33.46 \\
      \hline
      DGRC$_{(ft3,seq,shuf)}$ $^{\mathrm{b, c}}$& 58.14 & 37.75 & 25.19 & 20.07 & 23.38 & 32.67\\
      \hline
      DGRC$_{(ft3,e2e)}$ $^{\mathrm{b, c}}$& 61.32 & 41.06 & 28.31 & 21.08 & 29.56 & 38.35  \\ 
      \hline
      DGRC$_{(ft3,e2e,shuf)}$ $^{\mathrm{b, c}}$& \textbf{63.07 } & \textbf{42.08}  & \textbf{28.80}  & 20.98 & \textbf{29.85}  & \textbf{38.46}  \\ 
\hline
\multicolumn{7}{l}{$^{\mathrm{a}}$Q-aware = question-aware, A-aware = answer-aware, QA-aware = question-enhanced answer-aware.}\\
\multicolumn{7}{p{0.7\linewidth}}{$^{\mathrm{b}}$ft1 = question-aware fine-tuning, ft2 = answer-aware fine-tuning, ft3 = question-enhanced answer-aware fine-tuning.}\\
\multicolumn{7}{l}{$^{\mathrm{c}}$e2e = end-to-end mask pattern, seq = sequential mask pattern, shuf = shuffle mechanism.}
    \end{tabular}
\end{table*}

\section{Experiments}
\subsection{Data and Metrics}
We conducted experiments on \(C^3\) and Logiqa, comprising approximately 19.6K natural questions and 9K paragraph-question pairs from real examinations. 

After data cleaning, the training set consists of 5.7K templated natural questions and 9.1K non-templated questions, with a development set of 3K and a testing set of 3K. 
Through shuffling, we expanded the training set to almost 88.29K. 

We utilized BLEU \cite{papineni2002bleu}, ROUGE-L \cite{lin2004rouge}, and METEOR \cite{banerjee2005meteor} as automated evaluation metrics. 
BLEU gauges the average n-gram overlap among reference sentences, while METEOR and ROUGE-L adapt BLEU's n-gram overlap concept for machine translation and text summarization evaluation, respectively.  

\subsection{Implementation Details}
We employed the GLM large Chinese model with 335M parameters and set the maximum sequence length to 512.
The optimization was carried out using AdamW\cite{loshchilov2017decoupled}, with the learning rate set to 2e-5. 
Training was halted if the validation BLEU score did not show improvement for 8 epochs. 
We also applied gradient clipping with a length of 10.

\subsection{Main Results}
The experimental results for the testing set were evaluated using BLEU, METEOR, and ROUGE-L scores, as illustrated in Table~\ref{tab:exp}. 
DGRC demonstrates a more than 2.5-fold improvement in BLEU scores.   
And without fine-tuning, GLM and ChatGLM3 demonstrate superior performance, employing question-enhanced answer-aware prompting. 

\subsubsection{Comparing Fine-Tuning Strategies}
Experiments indicate that answer-aware fine-tuning may result in more effective prompts regarding BLEU-4 scores. 
Additionally, the shuffling mechanism significantly enhances model performance with finer granularity, enabling models to leverage provided inputs to generate distractors, as shown in DGRC$_{(ft3,e2e,shuf)}$. 

\subsubsection{Comparing Generation Mask Patterns}
Experiments indicate that the end-to-end mask pattern is a concise and effective approach to fine-tuning DG models. 
It achieves significant improvement in all metrics. 
Meanwhile, the sequential mask pattern incurs resource costs, and experimental results suggest that conditional generation does not enhance models' performance in automatic evaluation. 

\subsection{Ablation Study}
To further investigate the performance of different components and strategies within DGRC, we conducted an ablation study on the test set and presented the results in Table~\ref{tab:ablation study}. 

\subsubsection{Fine-Tuning w/o Hard CoT}
In this ablation study, we examined the effect of removing the hard CoT mechanism. 
The model was fine-tuned exclusively for multi-task learning.

\subsubsection{Fine-Tuning w/o ML}
This ablation study focused on the impact of multi-task learning. 
The model was fine-tuned in the distractor generation task and employing the hard CoT approach.

\subsubsection{Fine-Tuning w/o ML and CoT}
We investigated the impact of removing hard CoT and multi-task learning. 
The model underwent fine-tuning solely for distractor generation.

Firstly, our ablation study suggests that applying our fine-tuning framework for distractor generation markedly enhances performance in this task.  
Comparing the results between DGRC$_{(ft2,e2e)}$ and DGRC w/o fine-tuning reveals that our framework achieves a more than 2.5-fold improvement of BLEU scores.  

Secondly, experiments examining DGRC w/o hard CoT and DGRC w/o ML + hard CoT indicate that multi-task learning contributes to performance enhancement, resulting in an increase of 18.29\%  in BLEU scores.  

Thirdly, a significant increase of 1.92 BLEU points can be observed when fine-tuning our model with hard CoT, comparing DGRC w/o ML to DGRC w/o ML + hard CoT and DGRC$_{(ft2, e2e)}$ to DGRC w/o hard CoT. 

By incorporating hard CoT, the model can deduce the answer before generating distractors, thereby enhancing the quality of DG. 
However, the improvement from hard CoT is constrained because the templated questions are more than half the non-templated ones. 

\begin{table}[htbp]
\caption{The experiments results of ablation study}\label{tab:ablation study}
\begin{center}
\begin{tabular}{|l|c|}
\hline
 \textbf{Model}&\textbf{BLEU} \\
\hline
      DGRC w/o fine-tuning & 8.81\\
      \hline
      DGRC w/o ML + hard CoT & 18.04\\
      \hline
      DGRC w/o ML & 19.96\\
      \hline
      DGRC w/o hard CoT & 21.34\\
      \hline
      DGRC$_{(ft2,e2e)}$  & \textbf{22.64} \\
\hline
\end{tabular}
\end{center}
\end{table}

\subsection{Human Evaluation}
Metrics relying on n-grams for automatic evaluation might not holistically capture the quality of generated questions.  
To offer a more thorough assessment, we randomly selected 300 examples from the test set for human evaluation.  
This evaluation was performed on 300 randomly selected test samples, comprising 100 short (\textless 50 tokens), 100 medium (50-200 tokens), and 100 long (\textgreater 200 tokens) articles.  

Three independent evaluators were assigned the task of rating the 300 generated distractors on a scale from 1 (poor) to 5 (excellent) based on two criteria:
(1) Relevance: Assessing whether the distractors, while not answers, are pertinent to the passages and the questions; 
(2) Complexity: Determining whether the distractors can divert the reader's attention.  
The scores from the evaluators were averaged. 

Evaluation results indicate that models tend to produce correct options, particularly evident in models lacking fine-tuning.
After fine-tuning, the tendency to generate correct options is mitigated, and DGRC$_{(ft3,e2e,shuf)}$ emerges as the top performer. 
Although DGRC$_{(ft3,e2e,shuf)}$ does not attain the highest BLEU-4 scores, its superior performance in BLEU-1, BLEU-2, BLEU-3, METEOR, and ROUGE-L enhances the generation of high-quality distractors. 
Details are presented in Table~\ref{tab:human eval}. 

\begin{table*}[htbp]
\caption{The human evaluation for different models}\label{tab:human eval}
\begin{center}
\begin{tabular}{|l|c|c|c|c|c|c|c|c|}
\hline
\textbf{Model}
& \multicolumn{2}{|c|}{\textbf{Short Articles}} & \multicolumn{2}{|c|}{\textbf{Medium Articles} } & \multicolumn{2}{|c|}{\textbf{Long Articles} } & \multicolumn{2}{|c|}{\textbf{Average} } \\
\cline{2-9} 
\textbf{Name}& Rel. & Cpx. & Rel. & Cpx.  & Rel. & Cpx. & Rel. & Cpx\\
\hline
    ChatGLM3-6B & 3.34 &2.37 &4.10 &2.62 &3.94 &2.66 & 3.79 & 2.55\\
    \hline
     GLM & 3.02&2.03&3.02&2.03&3.79&2.03 & 3.28 & 2.03\\
      \hline
      DGRC$_{(ft2,e2e)}$ & 4.86 & 3.20 & 4.84 & 3.00 & 4.60 & 2.90 & 4.78 & 3.03\\
          \hline
      DGRC$_{(ft3,e2e,shuf)}$ & \textbf{4.89} & \textbf{3.33} & \textbf{4.88} & \textbf{3.35} & \textbf{4.87} & \textbf{3.45} & \textbf{4.87} & \textbf{3.38}\\
      \hline
      Ground-truth$^{*}$ & 5 & 3 & 5 & 3 & 5 & 3 & 5 &3\\
\hline
\multicolumn{9}{l}{*Evaluators compared the generated contents with the ground-truth.}\\

\end{tabular}
\end{center}
\end{table*}

\section{Conclusion}
The task of natural questions distractor generation presents unique challenges and merits thorough exploration, especially within the realm of Chinese multi-choice reading comprehension. 
To our knowledge, we are the first to delve into natural questions distractor generation in this context.

We introduce DGRC and hard CoT to address the challenges of natural question distractor generation in Chinese multi-choice reading comprehension. 
The novel framework DGRC integrates fine-tuning strategies, the hard CoT mechanism, multi-task learning, and generation mask patterns.

Experimental results show that DGRC greatly improves performance, and we hope our investigation will help researchers study the distractor generation task. 

\bibliography{DGRC.bib}{}
\bibliographystyle{IEEEtran.bst}

\end{document}